\documentclass{article}
\usepackage[preprint]{spconf}
\usepackage{amsmath,graphicx,amssymb}
\usepackage{url}

\copyrightnotice{\parbox[t]{\textwidth}{\footnotesize \copyright\ 2022 IEEE. Personal use of this material is permitted. Permission from IEEE must be obtained for all other uses, in any current or future media, including reprinting/republishing this material for advertising or promotional purposes, creating new collective works, for resale or redistribution to servers or lists, or reuse of any copyrighted component of this work in other works.}}

\graphicspath{{./figure/}}
\def \R {\mathbb{R}}
\def \x {\mathbf{x}}
\def \y {\mathbf{y}}
\def \LL {\mathcal{L}}

\title{SVG Vector Font Generation for Chinese Characters with Transformer}
\name{Haruka Aoki, Kiyoharu Aizawa}
\address{The University of Tokyo\\
Tokyo, Japan}

\begin{document}
\ninept
\maketitle
\begin{abstract}
Designing fonts for Chinese characters is highly labor-intensive and time-consuming.
While the latest methods successfully generate the English alphabet vector font, despite the high demand for automatic font generation, Chinese vector font generation has been an unsolved problem owing to its complex shape and numerous characters.
This study addressed the problem of automatically generating Chinese vector fonts from only a single style and content reference.
We proposed a novel network architecture with Transformer and loss functions to capture structural features without differentiable rendering.
Although the dataset range was still limited to the sans-serif family, we successfully generated the Chinese vector font for the first time using the proposed method.
  
\end{abstract}
\section{Introduction}
\label{sec:intro}
Designing Chinese vector fonts is generally a time-consuming and labor-intensive task, and thus there is a demand for automatic font generation to support creators.
While the generation of Chinese fonts as raster images \cite{azadi2018multi,zhang2018separating,10.1145/3355089.3356574,park2021mxfont} has been widely studied, the generation of fonts as vector graphics, which is the original form of fonts, has been barely tackled yet.
Scale-invariance of vector fonts is an essential property of fonts for use at various scales, from documents to posters, and for this reason, most of the fonts commonly used in the real world are in vector format.
Therefore, Chinese font generation in vector format is an attractive research problem and an essential task in computer graphics and computer vision.

SVG (Scalable Vector Graphics) \cite{svgw3c} is widely used as a standard format for representing vector graphics.
In SVG, a single graphic consists of paths, and each path represents an outline curve.
A path is a variable-length sequence of commands, and each command represents a parametric shape primitive.
Unlike conventional raster image generation, which generates luminance values of pixels lined up in a grid, vector graphic generation is similar to sequence generation in natural language processing (NLP).
However, one of the significant differences between vector graphic generation and NLP is that vector graphic representation is not unique.
Even if the rendered results have the same appearance, their command representations may vary, and this is one of the factors that make vector graphic generation challenging.

Another factor that makes vector graphic generation difficult is that it is hard to capture the characteristics of spatial arrangement and shape from a sequence of commands.
Rendering from vector graphics to raster images is commonly non-differentiable and cannot be incorporated into neural networks, so optimizing the appearance in raster image space is difficult.
Some recent studies \cite{Li2020Differentiable,wang2021deepvecfont} proposed differentiable rendering methods, but they are neither efficient in terms of computational cost nor accuracy, which makes it challenging to integrate them into a framework for better optimization.

In terms of vector font generation, early methods \cite{suveeranont2010example,Campbell2014Learning} were targeted at only alphabet fonts and required a lot of user intervention and topological constraints.
For Chinese vector fonts, Gao proposed a method \cite{gao2019automatic} to predict the layout by extracting parts from a large number of reference characters.

With the development of sequence generation methods using deep learning, several methods to generate a series of points using RNNs were proposed for sketches and Chinese character skeletons \cite{zhang2017drawing,ha2018a,tang2019fontrnn}.
Recently, methods have been proposed for generating outlines of alphabet fonts in SVG format with a more complex command system \cite{lopes2019learned,Alexandre2020deepsvg,wang2021deepvecfont}.
However, most of these vector font generation methods are for simple alphabet fonts, and they have not yet achieved the direct generation of Chinese character fonts with more complex shapes and numerous characters.

In this study, we tried to address this challenging problem: generating Chinese vector fonts from a single style reference.
Our contributions are as follows.
\begin{itemize}
  \item We identify a new problem of generating SVG Chinese characters from a style reference and content reference in a style-transfer manner.
  \item We propose a novel network that can generate complex Chinese characters using Transformer \cite{Vaswani2017Attention} and loss functions to capture spatial arrangement and shape. 
  \item We build a dataset consisting of 66 types of sans-serif fonts and 800 characters.
  \item We show for the first time, the proposed method directly generating Chinese vector fonts of sans-serif styles.
\end{itemize}


\section{Related Works}


Raster font generation has been widely studied \cite{azadi2018multi,zhang2018separating,10.1145/3355089.3356574,park2021mxfont}.
However, attempts to generate vector fonts have been limited, which we describe below.

Suveeranont and Igarashi \cite{suveeranont2010example} proposed a method that automatically extracts the outlines of reference characters and generates a new font by blending template fonts.
Campbell and Kautz \cite{Campbell2014Learning} learned the manifold of alphabet fonts and generated a new font by interpolating existing fonts.
For Chinese characters, Gao et al. \cite{gao2019automatic} proposed font generation as a layout prediction problem of the parts decomposed from 775 reference characters.

With the development of variable-length sequence generation methods using deep learning, some methods were proposed for predicting point sequences that form the skeleton of a character.
Zhang et al. \cite{zhang2017drawing} represented a Chinese character skeleton as a series of points (the points are connected by straight lines) and generated them using LSTM \cite{hochreiter1997long} or GRU \cite{cho2014learning}.
Tang et al. \cite{tang2019fontrnn} took a step further and proposed a method that simultaneously generated a raster image.

More recent methods generate character outlines.
SVG-VAE \cite{lopes2019learned} proposed by Lopes et al. trains a CNN-based VAE \cite{kingma2013auto} in raster domain and uses the extracted style embeddings to generate SVG that consists of straight lines and B\'{e}zier curves, with LSTM-based SVG decoder.
DeepSVG \cite{Alexandre2020deepsvg} proposed by Carlier et al. focuses on the hierarchical nature of SVG, that is image $>$ path $>$ command, and uses Transformer \cite{Vaswani2017Attention} hierarchically to encode/decode commands of vector graphics directly.
The overall architecture is a VAE that encodes style features, and the labels of the characters are input to both the encoder and decoder when generating fonts.
DeepSVG performs accurate reconstruction and interpolation of a simple icon dataset and generation of alphabet fonts.
DeepVecFont \cite{wang2021deepvecfont} by Wang et al. generates more visually pleasing alphabet fonts by simultaneously inputting and outputting raster and vector images using a shared feature space and optimizing outputs in two modalities.
It utilizes LSTM to process vector graphics. 
The output graphics are rasterized with a neural rasterizer to calculate the raster domain loss during training.
On the inference stage, the outputs are rasterized with a more accurate rasterizer \cite{Li2020Differentiable} for further refinement.

\section{Proposed Method}
\label{sec:prop}

\begin{table}[t]
  \caption{SVG drawing commands subset used for vector fonts. The start point $(x_0, y_0)$ is the end point of the previous command.}
  \label{tab:svg_cmds}
  \centering
  \begin{tabular}{ccc}
    \hline
    Commands & Arguments & Example \\
    \hline
    \parbox[c]{2.2cm}{\centering $\texttt{M}$\\(MoveTo)} & $x, y$ & \parbox[c]{3cm}{\vspace{1mm}\includegraphics[bb = 0 10 198 73,clip,width=\hsize]{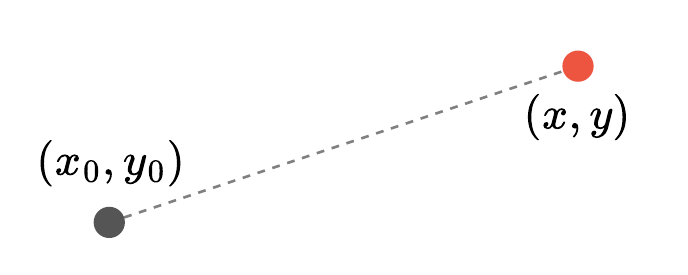}} \\
    \parbox[c]{2.2cm}{\centering $\texttt{L}$\\(LineTo)} & $x, y$ & \parbox[c]{3cm}{\includegraphics[bb = 0 0 198 65,clip,width=\hsize]{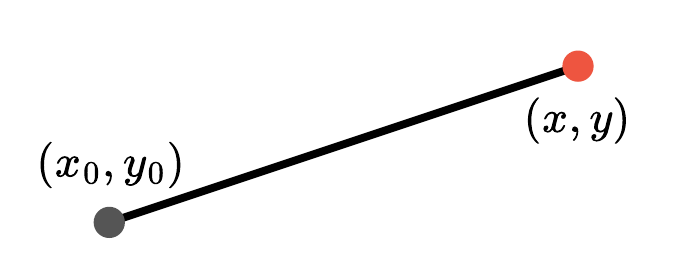}} \\
    \parbox[c]{2.2cm}{\centering $\texttt{C}$\\(Cubic B\'{e}zier Curve)} & \parbox[c]{1cm}{\centering $x_1, y_1,$ $x_2, y_2,$ $x, y$} & \parbox[c]{3cm}{\includegraphics[width=\hsize]{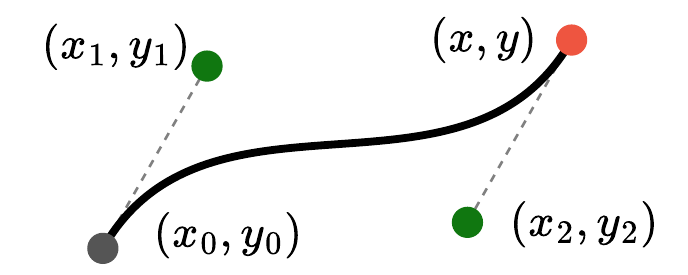}} \\
    \parbox[c]{2.2cm}{\centering $\texttt{Z}$\\(ClosePath)} & $\varnothing$ & \parbox[c]{3cm}{\includegraphics[width=\hsize]{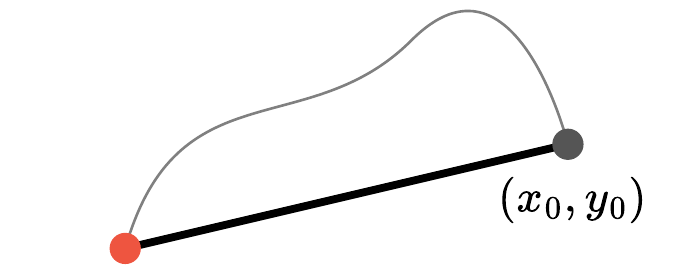}} \\
    \hline
  \end{tabular}
\end{table}

We propose a novel network architecture and loss functions to generate Chinese vector fonts from a content reference and a style reference in SVG format.
For SVG embedding, we use the same method as DeepSVG, which is described in Sec.~\ref{sec:datarep}.
The network architecture, which is the core of the proposed method, is described in Sec.~\ref{sec:network}.
The definition of the loss functions for optimization, including the proposed Chamfer loss and argument loss, is described in Sec.~\ref{sec:loss}.

\subsection{SVG Representation}
\label{sec:datarep}
To represent vector fonts, we use SVG \cite{svgw3c}, the standard format in various fields, such as Web graphics.
Among the various drawing commands in SVG, using the subset shown in Tab.~\ref{tab:svg_cmds} makes SVG interconvertible with OpenType fonts.

A single SVG graphic consisting of $N_P$ paths is denoted by $V = \{P_1, \dots, P_{N_P}\}$.
Each path is represented as $P_i = (S_i, v_i)$, with the visibility of the path $v_i \in \{0, 1\}$. We do not use the fill property used in DeepSVG.
Each $S_i = (C_{i}^1, \ldots, C_{i}^{N_C})$ consists of $N_C$ commands $C_i^j$.
The command is denoted by $C_i^j = (c^j_i, X_i^j)$, with the command type $c^j_i \in \{\texttt{<SOS>}, \texttt{M}, \texttt{L}, \texttt{C}, \texttt{Z}, \texttt{<EOS>}\}$ and its arguments.
$\texttt{<SOS>}, \texttt{<EOS>}$ represent the start and end tokens, respectively.
The list of arguments is the fixed-length expression $X_i^j = (x_1^{i,j}, y_1^{i,j}, x_2^{i,j}, y_2^{i,j}, x^{i,j}, y^{i,j}) \in \R^6$, where unused arguments are padded with $-1$.
We also treat the paths and commands as fixed-length sequences of $N_P$ and $N_C$, respectively, padded with $\texttt{<EOS>}$.

Each command $C_i^j$ is embedded into $e_i^j \in \R^{d_E}$.
$e_i^j $ is expressed as the sum of the command type embedding and the control point coordinates (arguments) in a similar manner as DeepSVG: $e_i^j = e_{\text{cmd}, i}^j + e_{\text{coord}, i}^j$.
The index embeddings is used in the first layer of networks.

\subsection{Network Architecture}
\label{sec:network}
\begin{figure}[tb]
  \begin{minipage}[b]{1.0\linewidth}
    \centering
    \centerline{\includegraphics[bb = 0 18 760 395,clip,width=0.86\hsize]{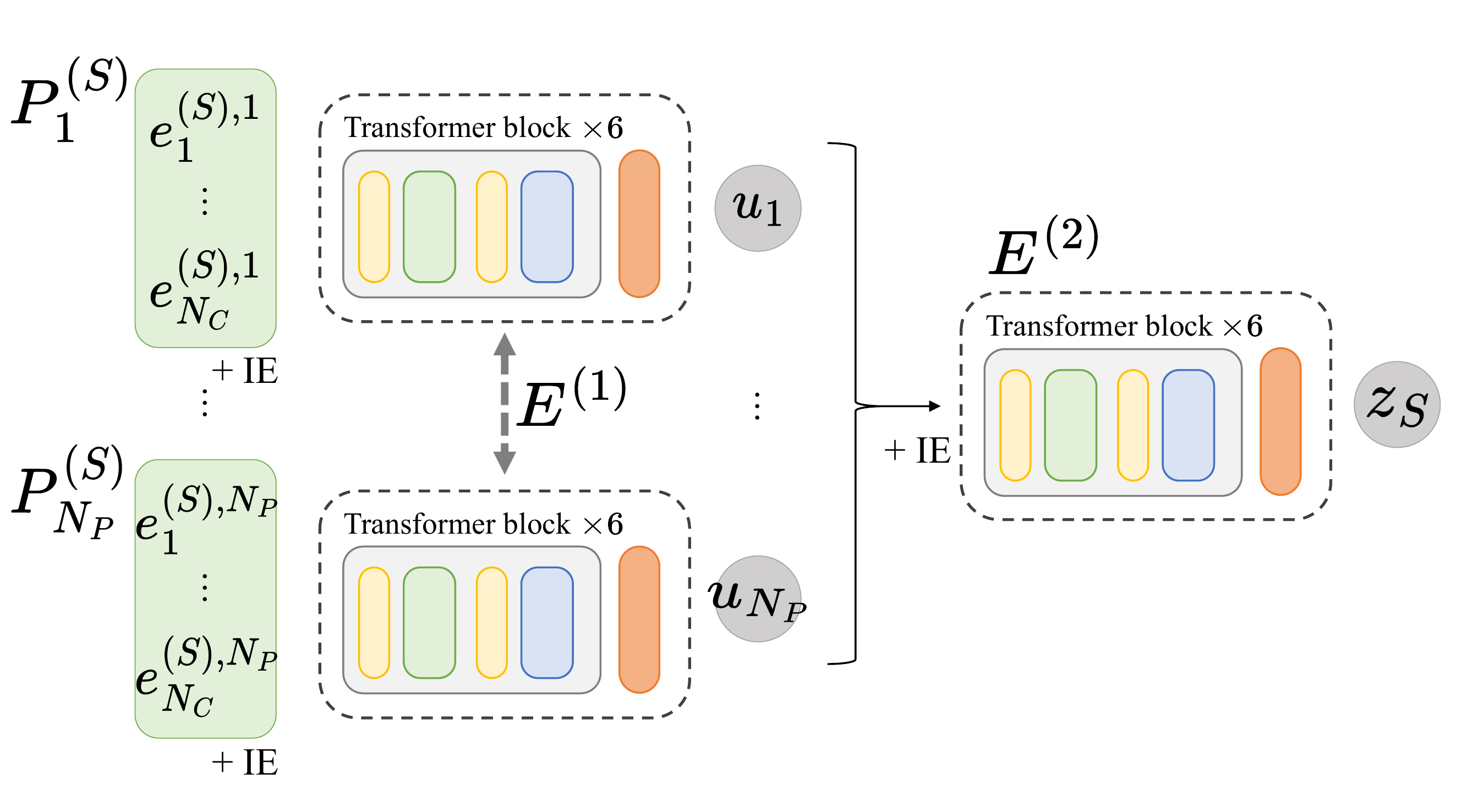}}
    \centerline{(a) Encoder.}\smallskip
    \label{fig:arch_e}
  \end{minipage}
  \begin{minipage}[b]{1.0\linewidth}
    \centering
    \centerline{\includegraphics[bb = 0 25 900 485,clip,width=0.97\hsize]{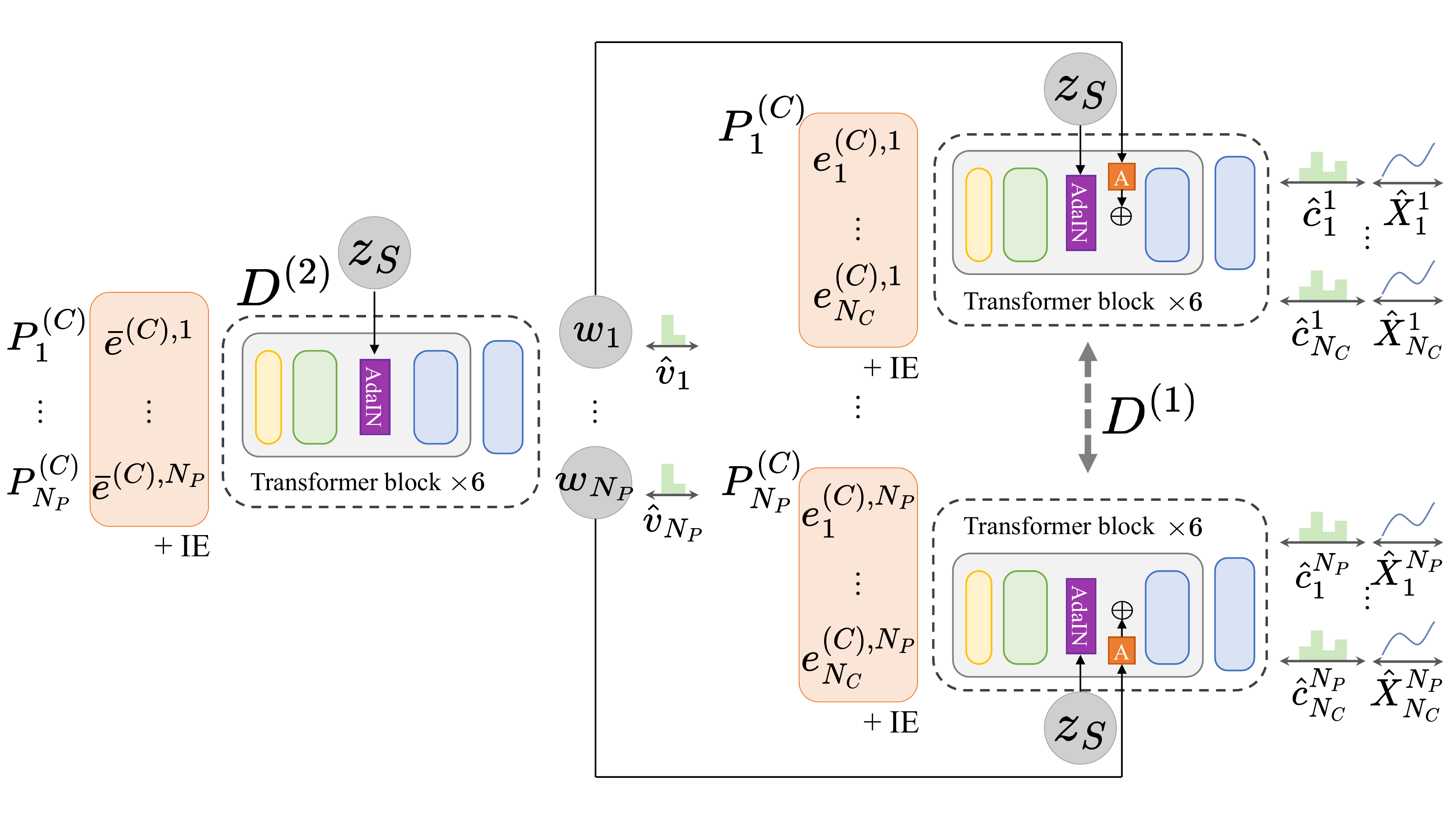}}
    \centerline{(b) Decoder.}\smallskip
    \label{fig:arch_d}
  \end{minipage}
  \caption{Network architecture. ``IE'' denotes index embedding.}
  \label{fig:arch}
\end{figure}

We propose a novel network architecture employing Transformer to generate SVG with the content feature of one content reference and style feature from a style reference; the overview is shown in Fig.~\ref{fig:arch}.
The encoder extracts the style feature $z_S$ from a style reference $V^{(S)} = \{P_1^{(S)}, \dots, P_{N_P}^{(S)}\}$.
The decoder generates the SVG output $\hat{V} = \{\hat{P}_1, \dots, \hat{P}_{N_P}\}$ from a content reference $V^{(C)} = \{P_1^{(C)}, \dots, P_{N_P}^{(C)}\}$ and $z_S$.
Like DeepSVG, they have a hierarchical structure consisting of $E^{(1)}$ and $D^{(1)}$, which process each path, and $E^{(2)}$ and $D^{(2)}$, which process the entire graphics.
The significant difference from DeepSVG is that the content is fed not as a label but a reference graphic, and, also, feeding style feature with AdaIN \cite{huang2017adain} on the decoder.
The detailed architecture of each module is shown in Fig.~\ref{fig:arch_detail}.
Each module has six Transformer blocks with an MLP dimension of 512 and an embedding dimension $d_E$ 256.
The prediction process is entirely feed-forward in both training and inference.

\subsubsection{Encoder}
The encoder extracts the style feature $z_S$ from the style reference $V^{(S)} = \{P_1^{(S)}, \dots, P_{N_P}^{(S)}\}$ (Fig.~\ref{fig:arch}(a)).
It uses the same structure as DeepSVG, except that it does not use VAE.
First, $E^{(1)}$ encodes each path independently.
For the path $P_i^{(S)}$, we input the embedding sequence $(e_j^{(S),i})_{j=1}^{N_C}$ to $E^{(1)}$ with learnable index embeddings, and obtain path feature $u_i$, which is the average-pooled outputs of the last layer.
Then the path-wise features $\{u_i\}_1^{N_P}$ are input to $E^{(2)}$ with index embeddings, and we obtained the average-pooled output $z_S$ as the style feature of the entire graphic.

\subsubsection{Decoder}

The decoder takes the content reference $V^{(C)} = \{P_1^{(C)}, \dots, P_{N_P}^{(C)}\}$ and the style feature $z_S$, and produces an output that reflects only the style of the latter while preserving the content of the former (Fig.~\ref{fig:arch}(b)).
For the content reference, we denoted the average of $(e^{(C),i}_j)_{j=1}^{N_C}$ by $\bar{e}^{(C),i}$, as path-wise embedding for the path $P_i^{(C)}$.
We added learnable index embeddings to $\{\bar{e}^{(C),i}\}_{i=1}^{N_P}$, and input it to $D^{(2)}$.
$D^{(2)}$ outputs path-wise feature representations $\{w_1, \dots, w_{N_P}\}$ and predictions of path visibility $\{\hat{v}_1, \dots, \hat{v}_{N_P}\}$.
Then, for each path in the content reference, we added index embeddings to $(e_j^{(C),i})_{j=1}^{N_C}$ and input it to $D^{(1)}$.
We appended the output $w_i$ of $D^{(2)}$ to the middle layer via a learnable affine transformation.
$D^{(1)}$ predicts the command type $\{(\hat{c}_{1}^i, \dots, \hat{c}_{N_C}^i)\}_{i=1}^{N_P}$ and the control point coordinates $\{(\hat{X}_{1}^i, \dots, \hat{X}_{N_C}^i)\}_{i=1}^{N_P}$ by MLP heads.
While the command type prediction is a classification problem, we directly predict continuous values within $[0, 255]$ for the coordinate values.

\begin{figure}[tb]
  \centering
  \includegraphics[bb = 0 9 709 444,clip,width=0.89\hsize]{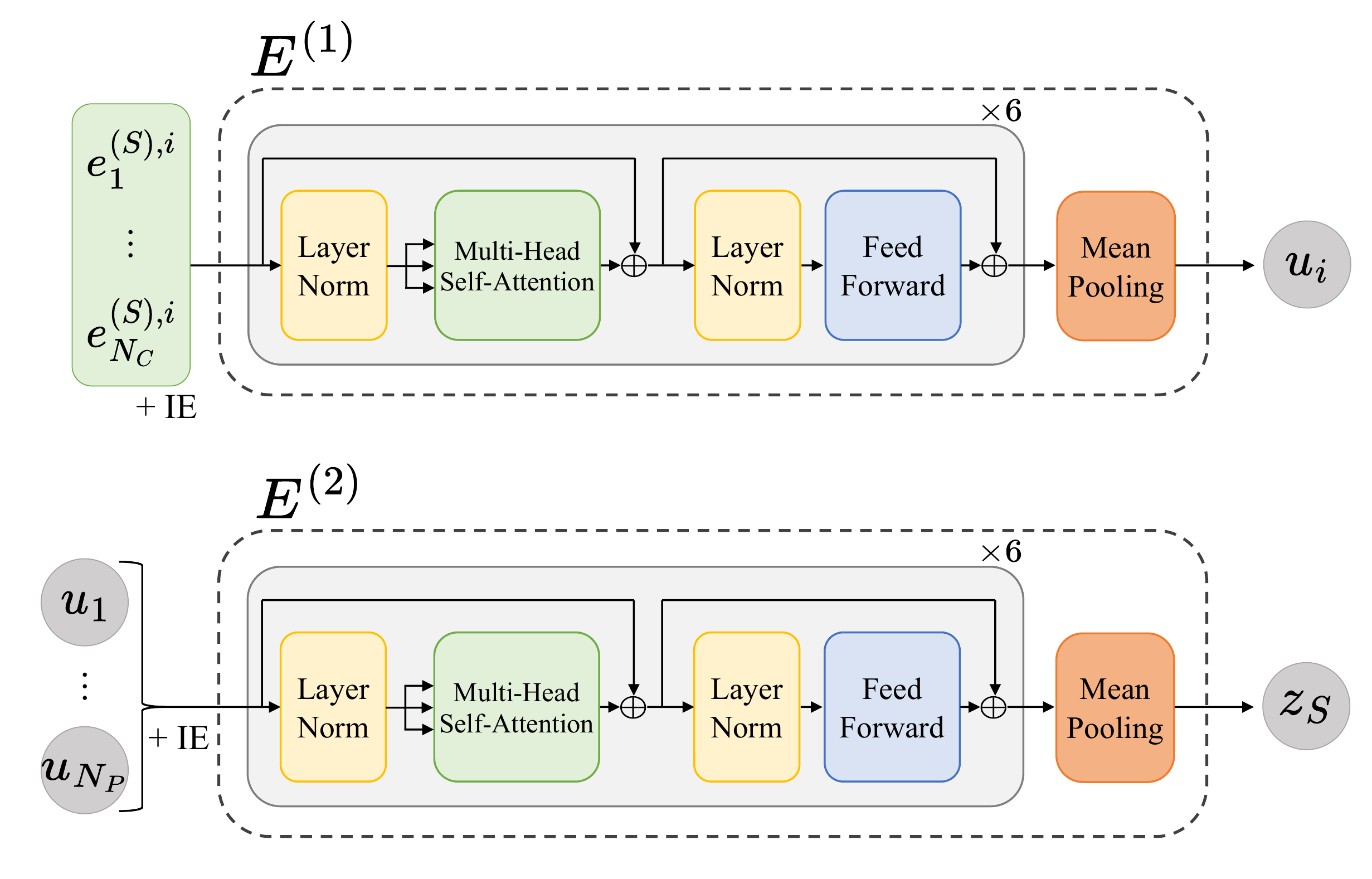}
  \\
  \includegraphics[bb = 0 20 950 610,clip,width=\hsize]{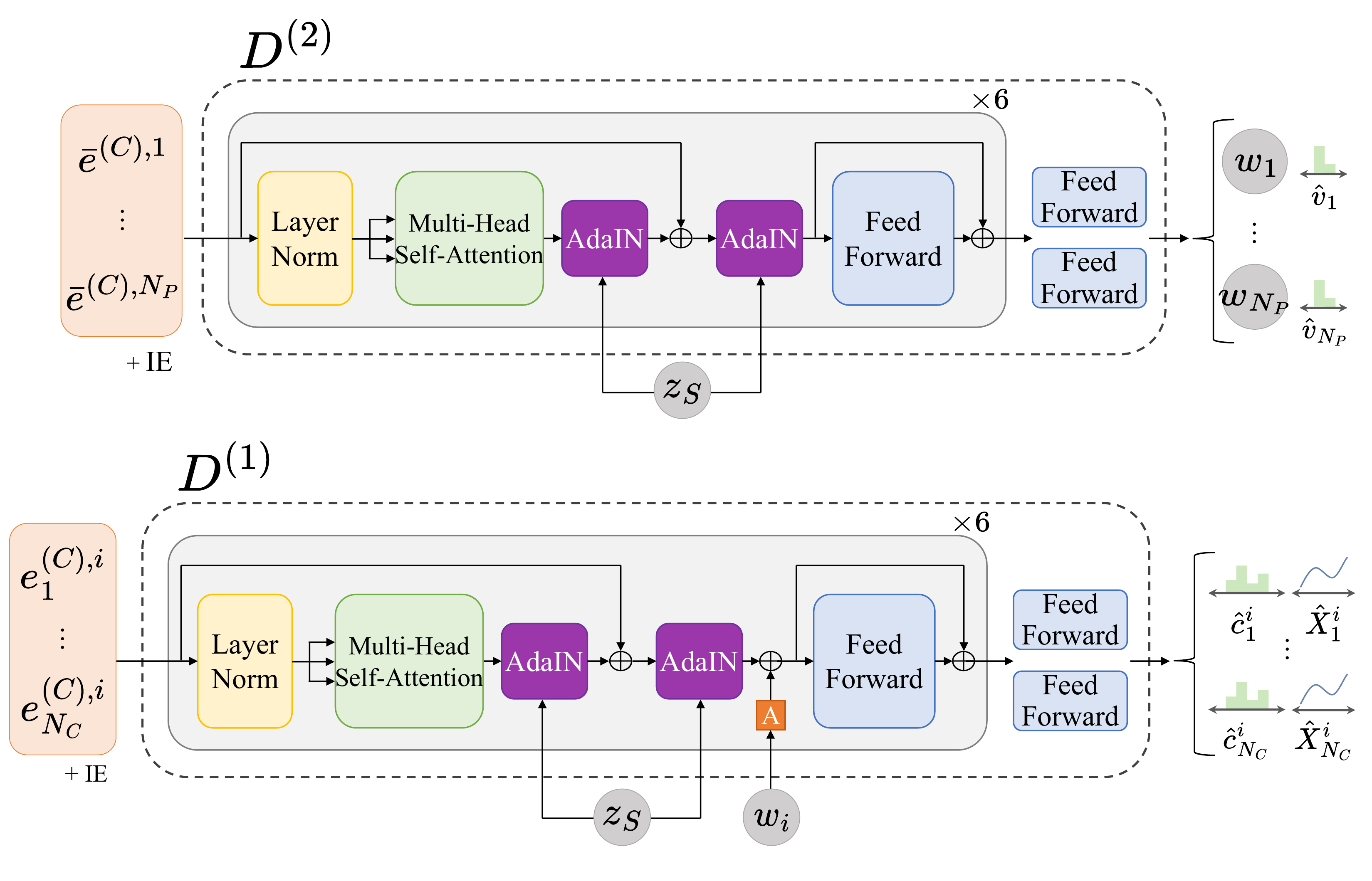}
  \caption{Detailed architecture of encoder and decoder.}
  \label{fig:arch_detail}
\end{figure}

To feed the style feature $z_S$, we used a method based on the config-c of STrans-G \cite{xu2021stransgan} (see Fig.~\ref{fig:arch_detail} for details).
Different from the original AdaIN \cite{huang2017adain} that performs style transfer between feature maps, AdaIN implemented by us acts on a global vector $z_S \in \R^{d_E}$, following Eq. \ref{eq:adain}, where the scale/shift parameter is transformed by MLP $\gamma(\cdot),\beta(\cdot)$.
\begin{equation}
    \text{AdaIN}(x, z_S) = \gamma(z_S) \left(\frac{x - \mu(x)}{\sigma(x)}\right) + \beta(z_S),
    \label{eq:adain}
\end{equation}
where $\mu(\cdot),\sigma(\cdot)$ denote the mean and standard deviation of the sequence, respectively.

\subsection{Loss Functions}
\label{sec:loss}

\begin{figure*}[tb]
  \begin{minipage}[b]{0.5\linewidth}
    \centering
    \centerline{\includegraphics[bb = 0 9 1440 720,clip,width=0.98\linewidth]{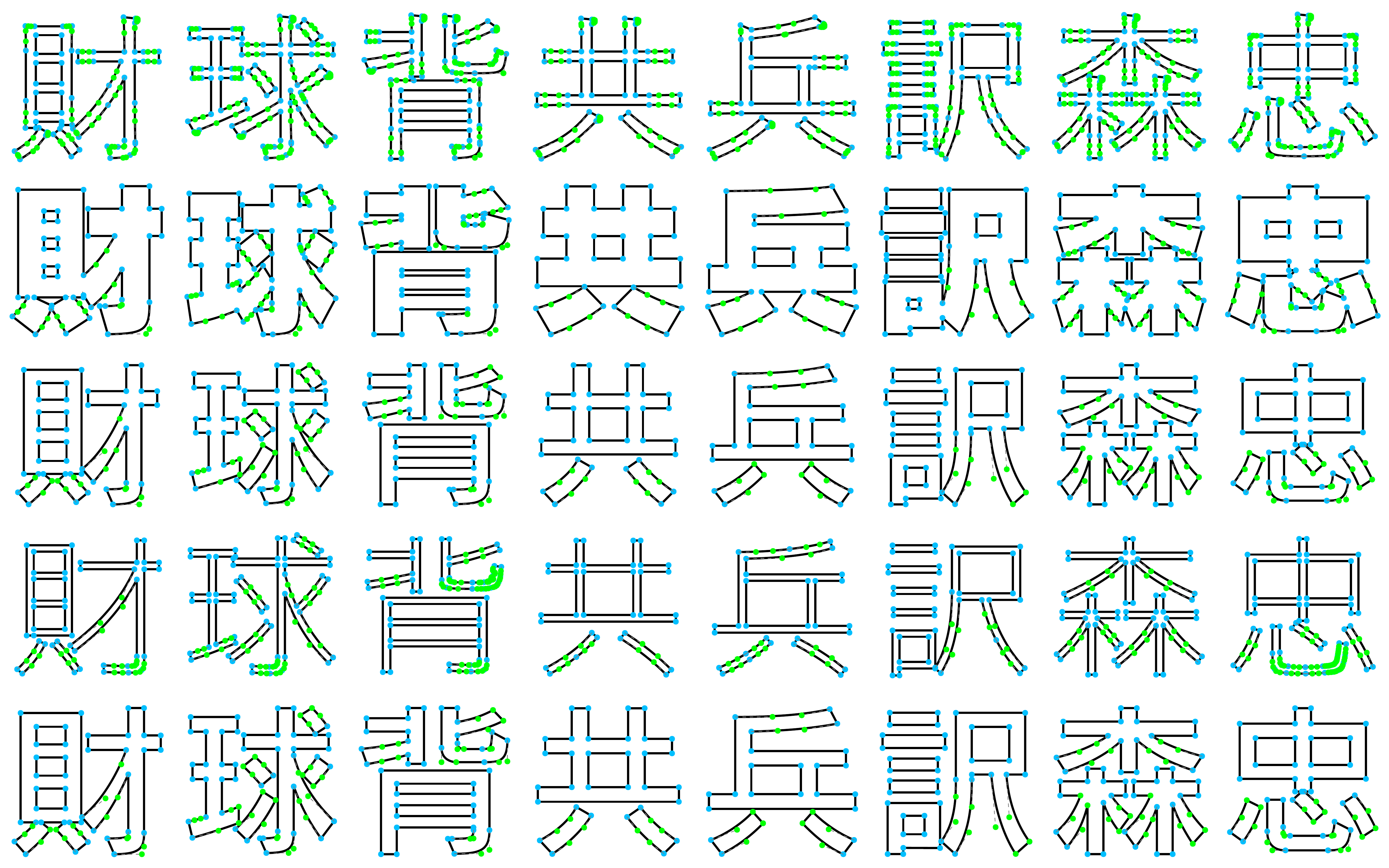}}
    \centerline{(a) Ground truth}\smallskip
    \label{fig:gt}
  \end{minipage}
  \begin{minipage}[b]{0.5\linewidth}
    \centering
    \centerline{\includegraphics[bb = 0 9 1440 720,clip,width=0.98\linewidth]{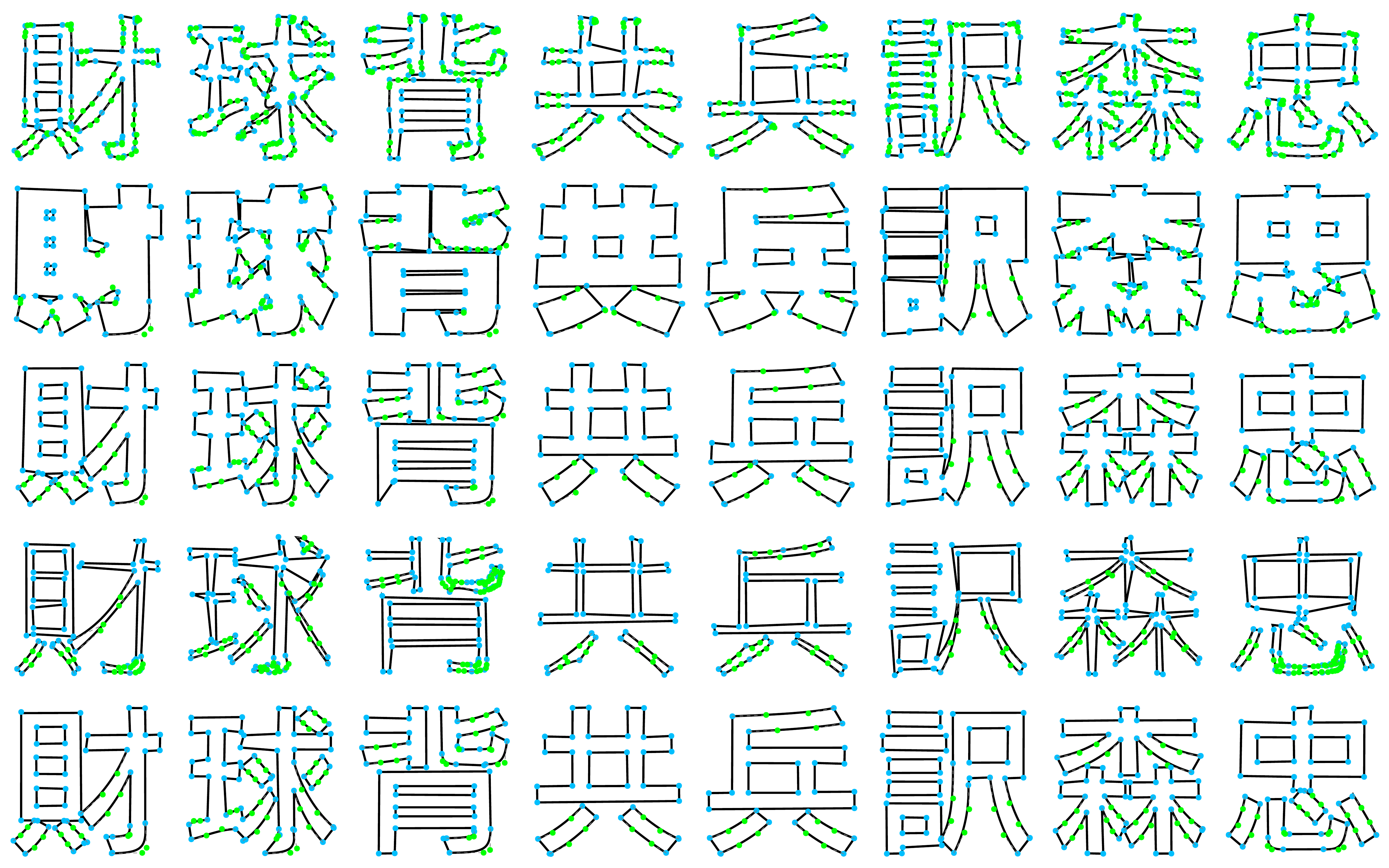}}
    \centerline{(b) Ours}\smallskip
    \label{fig:result_ours_full}
  \end{minipage}
  \begin{minipage}[b]{0.5\linewidth}
    \centering
    \centerline{\includegraphics[bb = 0 0 1440 720,clip,width=0.98\linewidth]{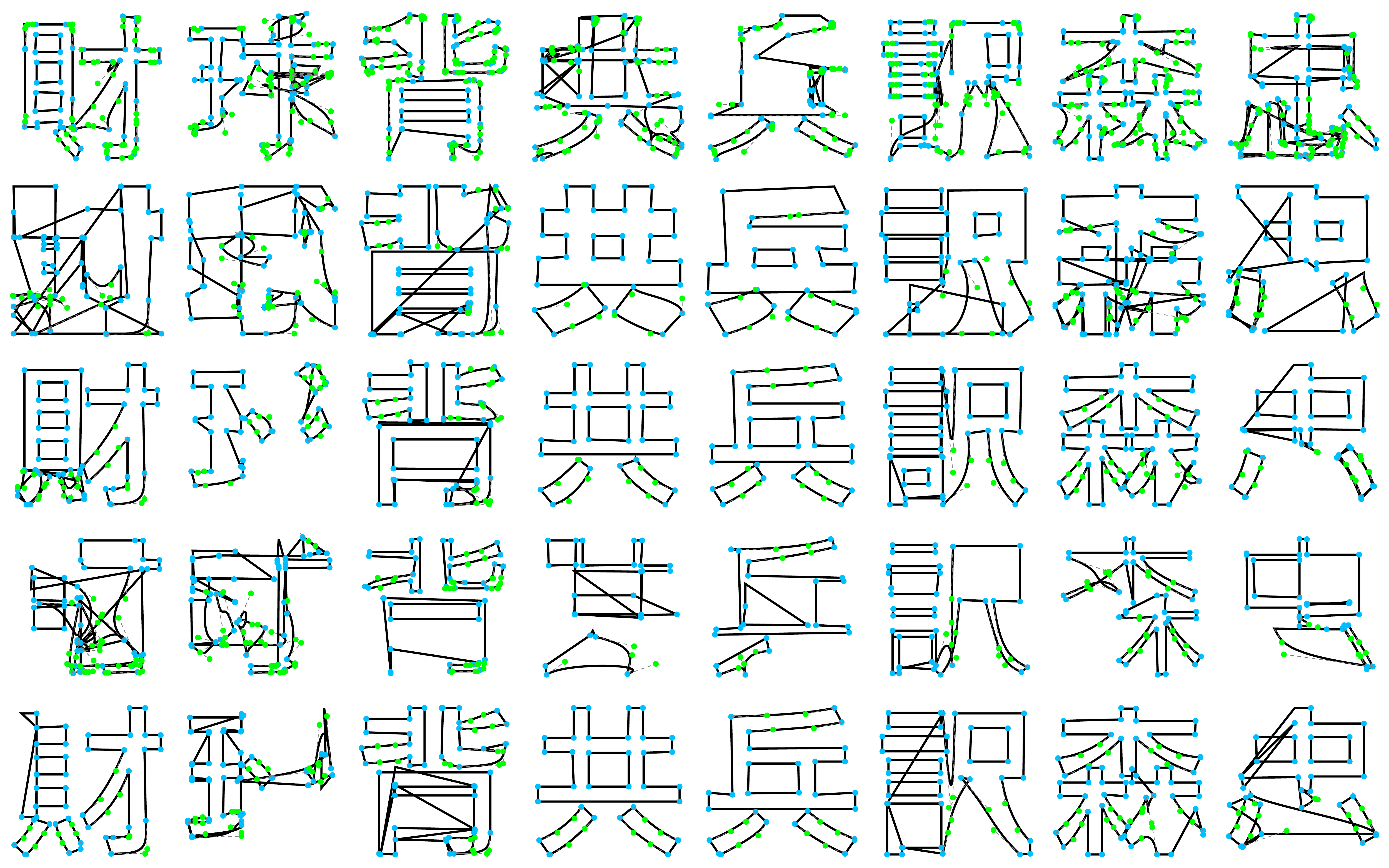}}
    \centerline{(c) DeepSVG \cite{Alexandre2020deepsvg}}\smallskip
    \label{fig:result_deepsvg}
  \end{minipage}
  \begin{minipage}[b]{0.5\linewidth}
    \centering
    \centerline{\includegraphics[bb = 0 0 1440 720,clip,width=0.98\linewidth]{vec_res/deepsvg_woaug.pdf}}
    \centerline{(d) DeepVecFont \cite{wang2021deepvecfont}}\smallskip
    \label{fig:result_dvf}
  \end{minipage}
  \caption{Generated vector fonts. Columns indicate content, Rows indicate style.}
  \label{fig:result}
\end{figure*}

The loss function consists of four terms: visibility loss, command-type loss, argument loss, and Chamfer loss.
Common to all, we define the correspondence between ground-truth paths and predicted paths by sorting the ground-truth paths according to a pre-defined order.
For visibility and command-type loss, we use cross-entropy loss.
For the argument loss for the control point coordinates, we calculated ${L_1}$ loss for each coordinate value, because our decoder outputted continuous values different from discrete values in DeepSVG.
Unused arguments in the ground truth were masked when calculating losses.
In addition, we propose to introduce Chamfer loss to capture spatial features without rasterizing predicted paths.
Among these four losses, Chamfer loss and argument loss are the new losses that we have introduced, the remaining two are the same as DeepSVG.
The details of Chamfer loss are described below.

Chamfer distance is a standard metric mainly in point cloud generation tasks \cite{Fan_2017_CVPR}, which is calculated between a pair of point clouds.
For two point clouds $Q_1, Q_2 \subseteq \R^2$, it is defined as
\begin{align}
  & d_{\text{CD}}(Q_1, Q_2) =  \nonumber \\
  &~~~~ \frac{1}{|Q_1|} \sum_{\x \in Q_1} \min_{\y \in Q_2} ||\x - \y||_2^2 + \frac{1}{|Q_2|} \sum_{\y \in Q_2} \min_{\x \in Q_1} ||\x - \y||_2^2 .
\end{align}
We apply this to calculate the distance between SVG paths.
First, we sample $n_p$ points on each curve defined by SVG command $C_i^j$, as

\begin{equation}
  \phi_{i,j} = 
  \begin{cases}
      \{\textbf{r}_{i,j} (k/n_p) \}_{k=0}^{n_p-1} & (c_i^j \in \{ \texttt{L},\texttt{C} \})\\
      \varnothing & (\text{else})
  \end{cases},
\end{equation}
where $\textbf{r}_{i,j}(t)$ $(0 \le t \le 1)$ is parametric representation of the curve $C_i^j$.
The point cloud $\Phi_i$ for the ground-truth path $P_i$ is defined by
\begin{equation}
    \Phi_i = \bigcup_j^{N_C} \phi_{i,j}.
\end{equation}
The point cloud for the predicted path $\hat{P}_i$ can be calculated similarly and denoted as $\hat{\Phi}_i$.
Calculating $d_{\text{CD}}$ for each pair of $(\Phi_i, \hat{\Phi}_i)$, the Chamfer loss is defined as
\begin{equation}
  \LL_{\text{cfr}} = \frac{1}{N_P} \sum_i^{N_P} d_{\text{CD}}(\Phi_i, \hat{\Phi}_i).
  \label{eq:cfr_loss}
\end{equation}
There is a trade-off between accuracy and computational cost, which can be adjusted by $n_p$.
We use $n_p=9$ during training.

\section{Experiment}

\subsection{Dataset}
We built a dataset of Chinese characters on Japanese fonts, which consisted of the products of Fontworks Inc. and free fonts from the Web, to evaluate the effectiveness of the proposed method.
Since it is quite challenging to generate various and complex Chinese characters, we limited the dataset to 66 sans-serif fonts (styles) and 800 relatively simple characters.
We randomly chose 59 fonts for training and the remaining seven for evaluation.
The font for content reference was fixed for the entire learning and inference process.

\subsection{Implementation Details}
We trained the model for 1500 epochs with a batch size of 96.
The learning rate started at 0.0, increased linearly to 0.002 over 500 iterations, and then decayed exponentially.
The maximum number of paths was $N_P = 12$, and the maximum number of commands per path was $N_C = 100$.
We also conducted experiments comparing our method to two state-of-the-art methods, DeepSVG \cite{Alexandre2020deepsvg} and DeepVecFont \cite{wang2021deepvecfont}.
For DeepSVG, we increased $N_P$, $N_C$, the number of Transformer blocks, training epochs, and the batch size from the original settings to our settings to support Chinese fonts\footnote{In the original implementation, they performed data augmentation, but the results were better without it, so we report the results w/o augmentation.}.
For DeepVecFont, we set the maximum sequence length to 250 and trained for 100 epochs.

\subsection{Results}

We show the fonts generated by the proposed method in Fig.~\ref{fig:result}.
Compared to the other methods, our results are consistently closer to the ground truth. 
State-of-the-art methods suffered from artifacts, and some of their results could not be recognized as characters.
On the other hand, the proposed method correctly captured the style and content features and successfully generated visually pleasing fonts.

Quantitative evaluations are summarized in Tab.~\ref{tab:vector}.
For rasterized pixel distance, we rasterized \footnote{The inner regions are filled with black, following the ``even-odd'' rule} the generated/ground-truth SVG and computed the $L_1$ distance between the rasterized images.
Chamfer distance is similar to Chamfer loss, but instead of calculating the distance for each pair of paths, the distance is calculated for  pairs in the entire graphics by grouping all paths on one graphic.
Additionally, we increased the number of sampled points per command, $n_p$, to 99 for accurate evaluation.
The proposed method remarkably outperformed the latest methods for vector graphic generation in terms of all metrics.

\begin{table}[t]
  \caption{Quantitative evaluation.}
  \label{tab:vector}
  \centering
  \begin{tabular}{lcc}
      \hline
      Method & \begin{tabular}{c}Rasterized\\Pixel Distance\end{tabular} & Chamfer Distance\\
      \hline
      DeepSVG \cite{Alexandre2020deepsvg}                & 0.150 & 7.20 \\
      DeepVecFont \cite{wang2021deepvecfont}             & 0.209 & 8.32 \\
      Ours                                              & \textbf{0.058} & \textbf{3.09} \\
      \hline
  \end{tabular}
\end{table}

\section{Conclusion}
This paper proposed a new style-transfer network architecture using hierarchical Transformers using AdaIN and loss functions, including Chamfer loss, to capture the spatial structure of vector graphics efficiently.
SOTA methods for vector graphic generation failed to generate complex vector fonts such as Chinese characters.
The proposed method enabled us to successfully generate Chinese vector fonts from a single style reference.
We also demonstrated the effectiveness of our method in generating Chinese fonts in the sans-serif styles through comparison with SOTA methods.


\vfill\pagebreak

\bibliographystyle{IEEEbib}
\bibliography{refs}

\begin{thebibliography}{10}

\bibitem{azadi2018multi}
Samaneh Azadi, Matthew Fisher, Vladimir~G Kim, Zhaowen Wang, Eli Shechtman, and
  Trevor Darrell,
\newblock ``Multi-content gan for few-shot font style transfer,''
\newblock in {\em Proc. IEEE CVPR}, 2018, pp. 7564--7573.

\bibitem{zhang2018separating}
Yexun Zhang, Ya~Zhang, and Wenbin Cai,
\newblock ``Separating style and content for generalized style transfer,''
\newblock in {\em Proc. IEEE CVPR}, 2018, pp. 8447--8455.

\bibitem{10.1145/3355089.3356574}
Yue Gao, Yuan Guo, Zhouhui Lian, Yingmin Tang, and Jianguo Xiao,
\newblock ``Artistic glyph image synthesis via one-stage few-shot learning,''
\newblock {\em ACM Trans. Graph.}, vol. 38, no. 6, pp. 1--12, Nov. 2019.

\bibitem{park2021mxfont}
Song Park, Sanghyuk Chun, Junbum Cha, Bado Lee, and Hyunjung Shim,
\newblock ``Multiple heads are better than one: Few-shot font generation with
  multiple localized experts,''
\newblock in {\em Proc. IEEE ICCV}, 2021, pp. 13900--13909.

\bibitem{svgw3c}
{W3C SVG Working Group},
\newblock ``Scalable vector graphics ({SVG}),''
  \url{https://www.w3.org/Graphics/SVG/}.

\bibitem{Li2020Differentiable}
Tzu-Mao Li, Michal Luk\'{a}\v{c}, Micha\"{e}l Gharbi, and Jonathan
  Ragan-Kelley,
\newblock ``Differentiable vector graphics rasterization for editing and
  learning,''
\newblock {\em ACM Trans. Graph.}, vol. 39, no. 6, pp. 1--15, Nov. 2020.

\bibitem{wang2021deepvecfont}
Yizhi Wang and Zhouhui Lian,
\newblock ``Deepvecfont: Synthesizing high-quality vector fonts via
  dual-modality learning,''
\newblock {\em ACM Trans. Graph.}, vol. 40, no. 6, pp. 1--15, Dec. 2021.

\bibitem{suveeranont2010example}
Rapee Suveeranont and Takeo Igarashi,
\newblock ``Example-based automatic font generation,''
\newblock in {\em International Symposium on Smart Graphics}, 2010, pp.
  127--138.

\bibitem{Campbell2014Learning}
Neill D.~F. Campbell and Jan Kautz,
\newblock ``Learning a manifold of fonts,''
\newblock {\em ACM Trans. Graph.}, vol. 33, no. 4, pp. 1--11, July 2014.

\bibitem{gao2019automatic}
Yichen Gao, Zhouhui Lian, Yingmin Tang, and Jianguo Xiao,
\newblock ``Automatic generation of chinese vector fonts via deep layout
  inferring,''
\newblock in {\em SIGGRAPH Asia 2019 Technical Briefs}, 2019, SA '19, pp.
  33--36.

\bibitem{zhang2017drawing}
Xu-Yao Zhang, Fei Yin, Yan-Ming Zhang, Cheng-Lin Liu, and Yoshua Bengio,
\newblock ``Drawing and recognizing chinese characters with recurrent neural
  network,''
\newblock {\em IEEE TPAMI}, vol. 40, no. 4, pp. 849--862, 2017.

\bibitem{ha2018a}
David Ha and Douglas Eck,
\newblock ``A neural representation of sketch drawings,''
\newblock in {\em Proc. ICLR}, 2018.

\bibitem{tang2019fontrnn}
Shusen Tang, Zeqing Xia, Zhouhui Lian, Yingmin Tang, and Jianguo Xiao,
\newblock ``{FontRNN: Generating Large-scale Chinese Fonts via Recurrent Neural
  Network},''
\newblock {\em Computer Graphics Forum}, pp. 567--577, 2019.

\bibitem{lopes2019learned}
Raphael~Gontijo Lopes, David Ha, Douglas Eck, and Jonathon Shlens,
\newblock ``A learned representation for scalable vector graphics,''
\newblock in {\em Proc. IEEE ICCV}, 2019, pp. 7930--7939.

\bibitem{Alexandre2020deepsvg}
Alexandre Carlier, Martin Danelljan, Alexandre Alahi, and Radu Timofte,
\newblock ``Deepsvg: A hierarchical generative network for vector graphics
  animation,''
\newblock in {\em NeurIPS}, 2020, vol.~33, pp. 16351--16361.

\bibitem{Vaswani2017Attention}
Ashish Vaswani, Noam Shazeer, Niki Parmar, Jakob Uszkoreit, Llion Jones,
  Aidan~N Gomez, Lukasz Kaiser, and Illia Polosukhin,
\newblock ``Attention is all you need,''
\newblock in {\em NeurIPS}, 2017, vol.~30, pp. 5998--6008.

\bibitem{hochreiter1997long}
Sepp Hochreiter and J{\"u}rgen Schmidhuber,
\newblock ``Long short-term memory,''
\newblock {\em Neural computation}, vol. 9, no. 8, pp. 1735--1780, 1997.

\bibitem{cho2014learning}
Kyunghyun Cho, Bart van Merri{\"e}nboer, Caglar Gulcehre, Dzmitry Bahdanau,
  Fethi Bougares, Holger Schwenk, and Yoshua Bengio,
\newblock ``Learning phrase representations using {RNN} encoder{--}decoder for
  statistical machine translation,''
\newblock in {\em Proc. EMNLP}, Oct. 2014, pp. 1724--1734.

\bibitem{kingma2013auto}
Diederik~P Kingma and Max Welling,
\newblock ``Auto-encoding variational bayes,'' arXiv preprint arXiv:1312.6114,
  2013.

\bibitem{huang2017adain}
Xun Huang and Serge Belongie,
\newblock ``Arbitrary style transfer in real-time with adaptive instance
  normalization,''
\newblock in {\em Proc. IEEE ICCV}, 2017, pp. 1501--1510.

\bibitem{xu2021stransgan}
Rui Xu, Xiangyu Xu, Kai Chen, Bolei Zhou, and Chen~Change Loy,
\newblock ``The nuts and bolts of adopting transformer in gans,'' arXiv
  preprint arXiv:2110.13107, 2021.

\bibitem{Fan_2017_CVPR}
Haoqiang Fan, Hao Su, and Leonidas~J. Guibas,
\newblock ``A point set generation network for 3d object reconstruction from a
  single image,''
\newblock in {\em Proc. IEEE CVPR}, July 2017, pp. 605--613.

\end{thebibliography}

\end{document}